\documentclass{ecai2014}
\usepackage{times}
\usepackage{graphicx}
\usepackage{latexsym}
\usepackage{amsfonts}
\usepackage{amsmath}
\usepackage{cases}
\usepackage{mathrsfs}
\usepackage{graphicx}
\usepackage{amssymb}

\begin{document}

\title{Comparison of Multi-agent and Single-agent Inverse Learning on a Simulated Soccer Example}

\author{Xiaomin Lin  \institute{University of Virginia, USA, email: xl5db@virginia.edu} \and Peter A. Beling  \institute{University of Virginia, USA, email: pb3a@virginia.edu} \and Randy Cogill\institute{IBM Research, Ireland, email: rlc9s@virginia.edu}  }

\maketitle
\bibliographystyle{ecai2014}

\begin{abstract}
We compare the performance of Inverse Reinforcement Learning (IRL) with the relative new model of Multi-agent Inverse Reinforcement Learning (MIRL). Before comparing the methods, we extend a published Bayesian IRL approach that is only applicable to the case where the reward is only state dependent to a general one capable of tackling the case where the reward depends on both state and action. Comparison between IRL and MIRL is made in the context of an abstract soccer game, using both  a  game model in which the reward depends only on state and one in which it depends on both state and action.  Results suggest that the IRL approach performs much worse than the MIRL approach. We speculate that the underperformance of IRL is because it fails to capture equilibrium information in the manner possible in MIRL. 
\end{abstract}

\section{INTRODUCTION}
The Multi-agent Reinforcement Learning (MRL) problem was first proposed by proposed by Littman \cite{Littman1994} to address the  limiting assumption in Reinforcement Learning (RL) that potentially responsive agents in a system area part of a passive environment. In a RL model, an agent can fully control the state transition process by taking actions on its own (though some stochastic variation is allowed); in a MRL problem, by contrast, the state transition process is determined by joint actions of all interacting rational agents. This essential difference complicates the MRL problems. As pointed out by Hu and Wellman \cite{Hu1998}, the concept of optimality, which is explicitly defined in IRL problems, loses its meaning in MRL problems since any agent's payoff depends on others' choices of action.  In the absence of optimality, one can adopt as a solution concept the \emph{Nash equilibrium}, in which each agent's choice is the best response to other agents' choices \cite{Hu1998}. In fact, so far there is no agreement on a solution concept for a general MRL problem. 
\par
The first attempt to solve a MRL problem, given by Littman \cite{Littman1994}, made use of a Markov or stochastic game \cite{Owen1968}, which is an extension of game theory to \emph{Markov Decision Process} (MDP)-like environments. However, only the special case of \emph{two-player zero-sum games}, in which one agent's gain is always the other's loss, is considered in \cite{Littman1994}. Hu and Wellman \cite{Hu1998} extended Littman's work, proposing a \emph{two-player general-sum} stochastic game framework for the MRL problem. Later MRL work has focused on the development of solution concepts and methods. For example, in \cite{Abdallah2008} a weak condition where an agent can neither observe other agents' actions or rewards, nor knows the underlying game or the corresponding Nash equilibrium a priori is considered and a new MRL algorithm called the Weighted Policy Learner (WPL) is proposed. Multi-agent learning in complex large distributed systems is also touched in \cite{Kash2011}, where it is noted that, although sophisticated multi-agent learning algorithms generally do not scale, it is possible to find restricted classes of games where simple efficient algorithms converge.  Solution concepts for distributed, multi-agent planning problems that involve coordination games under weak information exchange models have been considered in \cite{Patek2007,Zhao2008}.
\par
Inverse Reinforcement Learning (IRL), as the inverse learning problem for RL, has been studied extensively \cite{Ng2000,Qiao2011,Levine2011,Krishnamurthy2010,Ramachandran2007}. IRL aims to recover the reward function of the agent, in order to reason its behavior that is observed. Similarly, the inverse learning problems for MRL, termed MIRL, includes the problem of estimating the game payoffs being played, given only observations of the actions taken by the players. The reward function of an agent of interest recovered from an IRL approach, is in fact the mathematical expectation of all reward functions of other adaptive agents, which can be recovered from a MIRL approach. For example, if there are $n$ adaptive agents in the environment and the one of interest is the $k$th of them, its reward at state $s$ is 
\begin{equation*}
r^k\left ( s, a^k \right )=\iint_{A^k}r\left ( s, A^k \right )P\left ( A^k|s \right )dA^k,
\end{equation*}
where 
\begin{equation*}
\begin{aligned}
 A^k &= a^1\cdots a^{k-1}a^{k+1}\cdots a^n \\
 P\left ( A^k|s \right ) &= p\left ( a^1|s \right )\cdots p\left ( a^{k-1}|s \right )p\left ( a^{k+1}|s \right )\cdots p\left ( a^n|s \right ).
\end{aligned}
\end{equation*}
The above equation is valid in a general situation where the action space is continuous. 

There exist several solution approaches for MIRL. Natarajan \cite{Natarajan2010} presents an inverse reinforcement learning approach for multiple agents. However, that approach neither deals with competing agents nor considers a game-theoretic model. In \cite{Waugh2011}, a form of the inverse equilibrium problem is discussed. However, that paper considers simultaneous one-stage games, rather than the sequential stochastic games.  We have recently developed  a Bayesian formulation for two-person zero-sum MIRL problems, in which an abstract soccer game, as a numerical example, is solved \cite{Beling2013,Lin2014}. 
\par
Recall that MRL was proposed because the state transition dynamics remains unknown if other adaptive agent's actions are not taken into account, so that IRL is difficult to implement. However, in the two-person zero-sum MIRL model presented in \cite{Beling2013}, it is assumed that the two agents' polices of actions over all states, are known or observed. Therefore, a complete state transition matrix can be obtained. Then a question is naturally raised: is a MDP-based IRL approach able to solve the rewards recovery problem in a multi-agent environment? This question is worth investigating deeply because if the answer is yes, there is no meaning to put addition effort into MIRL research.  
\par
The primary contribution of this work is to answer the above question. We first extend a Bayesian IRL approach, the idea of which was original proposed in \cite{Qiao2011}, to infer the unknown rewards. Then we apply it on an abstract soccer game example, comparing the results with those obtained from the MIRL approach introduced in \cite{Beling2013}. We consider two cases: one is that the unknown reward is only state dependent and the other We demonstrate that the results are much worse than that obtained from the MIRL approach introduced in \cite{Beling2013}. This finding gives us an in-depth understanding the fundamentals of MIRL and substantiate the value of research on this topic.
\par
The rest of the paper is structured as follows: Section 2 gives notation and concepts preliminary to MIRL, which is developed in the two-person zero-sum case in Section 3. Section 4 extends the Bayesian IRL approach proposed in \cite{Qiao2011} to a general form. Section 5 presents the soccer example and experiments in which the two methods are used to recover rewards.  Section 6 provides an evaluation of the quality of the learned rewards in terms of game playing success in simulations of the soccer game. Finally, Section 7 offers concluding remarks. 

\section{PRELIMINARIES}\label{section_2}
A finite state two-person \emph{discounted stochastic game} can be specified in terms of the state space $\mathcal{S}=\left \{ 1,2,\cdots ,N \right \}$, the action spaces $\mathcal{A}_1=\mathcal{A}_2=\left \{ 1,2,\cdots ,M \right \}$, reward functions ${r}^k: \mathcal{S}\times \mathcal{A}_1\times \mathcal{A}_2 \mapsto  \mathbb{R}$ for each player $k \in \left \{ 1,2 \right \}$, transition probabilities $p\left ( s'|s, a^1, a^2 \right )$, and a discount factor $\gamma \in \left [ 0, 1 \right )$. 
\par
In the MIRL model presented in \cite{Beling2013}, there are several rules governing the games played between two agents: First, each player has perfect knowledge of the other's rewards. Second, each player acts simultaneously in any state transition process and receives a reward depending on the starting state and their immediate actions. Third, they repeat the game over an infinite time horizon, aiming to accumulate maximum discounted rewards. In addition, all single-state games are zero-sum, which is, more specifically, $r^1\left ( s, a^1, a^2 \right ) = -r^2\left ( s, a^1, a^2 \right )$. Due to the symmetry in rewards between the two players, we can simply use $r$ to denote $r^1$. Also, it is assumed that the \emph{bipolicy} $\pi$ of the two competing agents, which is a collection of both agents' policies of actions over all states, denoted $\left ( \pi^1, \pi^2 \right ) $, is known.
\par
The bipolicy-dependent, discounted expected sum of rewards of player $1$ as a function of the initial state, which is known as the \emph{value function}, can be formulated as:
\begin{equation}\label{V_orginial}
V_{\pi}\left ( s \right ) = \sum_{t=0}^{\infty }\gamma^tE\left ( \tilde{r}_{\pi}\left ( s_t \right ) |s_0=s \right ),
\end{equation}
where $s_t$ denotes the state of the game at stage $t$. $\tilde{r}_{\pi}\left ( s_t \right )$ is the single-stage expected reward earned by player $1$ at state $s$ under bipolicy $\pi$, specifically, 
\begin{equation}\label{R_average}
\begin{aligned}
\tilde{r}_{\pi}\left ( s \right ) &=  \sum_{a^1, a^2}\pi^1\left ( s, a^1 \right )\pi^2\left ( s, a^2 \right )r\left ( s, a^1, a^2 \right ) \\
&= \left [ \pi^1\left ( s \right ) \right ]^Tr\left ( s \right )\pi^2\left ( s \right ).
\end{aligned}
\end{equation}
Let $r$ be the rewards vector of player $1$, whose length is $M^2N$. $\tilde{r}_{\pi}$ can be expressed as
\begin{equation}\label{r_ave}
\tilde{r}_{\pi}=B_{\pi}r.
\end{equation}
More details about $B_{\pi}$ can be found in \cite{Lin2014}.   
\par
The state transition probability matrix under bipolicy $\pi$, $G_{\pi}$, is a $N \times N$ matrix with elements specified as 
\begin{equation}\label{G_def}
g_{\pi}\left ( s' | s \right )=\sum_{a^1, a^2}\pi^1\left ( s, a^1 \right )\pi^2\left ( s, a^2 \right )p\left ( s'|s, a^1, a^2 \right ).
\end{equation}
\par
A significant concept in two-person zero-sum MIRL is \emph{minimax} bipolicy in which each player minimizes his own maximum loss.  This  is an equilibrium in that it has the property that neither player can change the game value in their favor given that the other player holds their policy fixed. In a two-person zero-sum stochastic game, we say that the two agents reach a minimax bipolicy if both of them employ a minimax strategy in every single-stage game.  
\section{BAYESIAN MIRL} \label{section_3}
In \cite{Beling2013}, the authors point out that the core of two-person zero-sum MIRL is the assumption that two agents reach a minimax bipolicy because both of them are rational. Using all terminologies and notations introduced in the preceding section, a convex quadratic program, in a Bayesian optimization setting, can be proposed for a general two-person zero-sum MIRL problem
\begin{equation}\label{complex_program}
\begin{aligned}
\textrm{minimize:}  \quad
& \frac{1}{2}\left ( {r-{\mu_r}} \right )^T {\Sigma_r}^{-1} \left ( {r-{\mu_r}} \right )   \\
\textrm{subject to:} \quad
&\left ( B_{{\pi}^2|a^1=i}-B_{{\pi}} \right )D_{{\pi}}{r}\leq {0}\\
\quad
&\left ( B_{{\pi}^1|a^2=j}-B_{{\pi}} \right )D_{{\pi}}{r}\geq {0} \\
\end{aligned}
\end{equation}
for all $i\in\mathcal{A}_1$ and $j\in\mathcal{A}_2$, where $\mu_r$ is the mean of $r$ and $\Sigma_{{r}}$ is its covariance matrix. In the constraints of \eqref{complex_program}, $B_{{\pi}^{k}|a^{(3-k)}=i}$ ($k=1,2$) is conceptually similar to $B_{\pi}$, except that $B_{{\pi}^{(3-k)}|a^{k}=l}$ is constructed from a bipolicy in which player $k$ always takes action $l$ in any state while the other player still follows its original policy $\pi^{(3-k)}$. In addition, $D_{\pi}$ can be expanded as 
\begin{equation}\label{D_def}
D_{{\pi}} = \left ( I + \gamma P\left ( I - \gamma G_{{\pi}} \right )^{-1}B_{{\pi}} \right ), 
\end{equation}
where $P$ is a $NM^2 \times N$ matrix with $p\left ( s'|s, a^1, a^2 \right )$ as its elements.
\par
When it is known that $r$ is only a function of state, we can use a simper version of \eqref{complex_program}, as follows
\begin{equation}\label{simple_program}
\begin{aligned}
\textrm{minimize:}  \quad
& \frac{1}{2}\left ( {r}-\mu_{{r}} \right )^T \Sigma_{{r}}^{-1}\left ( {r}-\mu_{{r}} \right )  \\
\textrm{subject to:}  \quad
&\left ( G_{{\pi}}-G_{{\pi}^2|a^1=i} \right )\left ( {I}-\gamma G_{{\pi}} \right )^{-1}{r} \geq {0} \\
                          &\left ( G_{{\pi}}- G_{{\pi}^1|a^2=j} \right )\left ( {I}-\gamma G_{{\pi}} \right )^{-1}{r} \leq {0} \\
\end{aligned}
\end{equation}
for all $i\in\mathcal{A}_1$ and $j\in\mathcal{A}_2$, where $G_{{\pi}^{(3-k)}|a^k=l}$ ($k=1,2$) has a similar definition to that of $G_{\pi}$, except that $G_{{\pi}^{(3-k)}|a^k=l}$ is such a $N \times N$ state transition matrix that player $k$ always takes action $l$ in any state while the other player still follows its original policy $\pi^{(3-k)}$, the elements of which are, more specifically, 
\begin{equation}
g_{{\pi}^{(3-k)}|a^k=l}=\sum_{a^{(3-k)}}\pi^k\left ( s, l \right )\pi^{(3-k)}\left ( s,a^{(3-k)} \right )p\left ( s'|s, l, a^{(3-k)} \right ).
\end{equation}
\par
The theoretical validation of \eqref{complex_program} and \eqref{simple_program} are detailed in \cite{Beling2013,Lin2014}.
\section{BAYESIAN IRL} \label{sec_IRL}
We will address the multi-agent inverse problem from the perspective of IRL. Qiao and Beling \cite{Qiao2011} propose a Bayesian optimization program based on the assumption that the agent's reward function is only state dependent. In this section, we will extend this idea and formulate a more general program where the case that reward is state and action dependent is considered. Although we now turn to the MDP framework, most of the terminologies and notations introduced in Section \ref{section_2} will still be adopted here, unless otherwise specified. 
\par
As stated before, we will focus on player $1$'s rewards. However, we are now tasked to recover $r_{\pi^2}\left ( s, a^1 \right )$, which is the expected value of $r\left ( s, a^1, \pi^2\left ( s \right ) \right )$ in case player $2$ employs policy $\pi^2$, specifically, 
\begin{equation} \label{reward_transform}
r_{\pi^2}\left ( s, a^1 \right ) = \sum_{a^2}r\left ( s, a^1, a^2 \right )\pi^2\left ( s, a^2 \right ).
\end{equation}
\par
For simplicity, we will just use $r$ to denote the column vector whose element is $r_{\pi^2}\left ( s, a^1 \right )$, as follows:
\begin{equation*}
\begin{aligned}
r_{\pi^2}=(&\underbrace{r_{\pi^2}\left ( s_1, a_1^1 \right ),r_{\pi^2}\left ( s_2, a_1^1 \right ),\cdots ,r_{\pi^2}\left ( s_N, a_1^1 \right )}_{r_{a_1^1}},\cdots , \\
&\underbrace{r\left ( s_1, a_M^1 \right ),r_{\pi^2}\left ( s_2, a_M^1 \right ),\cdots ,r_{\pi^2}\left ( s_N, a_M^1 \right )}_{r_{a_M^1}})^T.
\end{aligned}
\end{equation*} 
Note that the length of $r$ here is $MN$, which is different from the one defined in Sections \ref{section_2} and \ref{section_3}. 
\par
We define player $1$'s Q-function of state $s$ and action $a^1$ under policy $\pi^1$, $Q_{\pi^1}\left ( s,a^1\right )$, to be the expected return from state $s$, taking action $a^1$ and thereafter following its original policy. 
\begin{equation}\label{Q_element_IRL}
Q_{\pi^1}\left ( s,a^1\right )=r_{\pi^2}\left ( s, a^1 \right ) + \gamma \sum_{s'}p\left ( s'|s, a^1 \right )V_{\pi^1}\left ( s' \right ),
\end{equation}
and its value function in state $s$ is 
\begin{equation}\label{V_IRL}
\begin{aligned}
V_{\pi^1}\left ( s \right ) &= \sum_{a^1}Q_{\pi^1}\left ( s,a^1\right )\pi^1\left ( s, a^1 \right ) \\
& = \tilde{r}_{\pi^1}\left ( s \right ) + \gamma\sum_{s'}g_{\pi}\left ( s'|s \right )V_{\pi^1}\left ( s' \right ),
\end{aligned} 
\end{equation}
where 
\begin{equation} \label{r_s_ave}
\tilde{r}_{\pi^1}\left ( s \right ) = \sum_{a^1}r_{\pi^2}\left ( s, a^1 \right )\pi^1\left ( s, a^1 \right ).
\end{equation}
Hence \eqref{V_IRL} can be written in matrix notation as
\begin{equation}\label{V_comp_IRL}
V_{\pi^1}=\tilde{r}_{\pi^1}+\gamma G_{\pi}V_{\pi^1},
\end{equation}
where 
\begin{equation}\label{r_ave_IRL}
\tilde{r}_{\pi^1}=C_{\pi^1}r,
\end{equation}
and $C_{\pi^1}$ is a $N \times NM$ matrix constructed from $\pi^1$, whose $i$th row is,
\begin{equation*}
\left [ \underbrace{0, \cdots ,0,\pi^1\left ( i, 1 \right ), 0,\cdots ,0}_{N},\underbrace{\cdots}_{\left ( M-2 \right )N} , \underbrace{0, \cdots ,0,\pi^1\left ( i, M \right ), 0,\cdots ,0}_{N}\right ].
\end{equation*}
Thus
\begin{equation}\label{V_comp2}
V_{\pi^1}=\left ( I - \gamma G_{\pi} \right )^{-1}C_{\pi^1}r.
\end{equation} 
The policy $\pi^1$ is optimal for player $1$ in the sense that 
\begin{equation}\label{fundamental}
V_{\pi^1}\left ( s,\pi^1\left ( s \right ) \right )\geqslant Q_{\pi^1}\left ( s,i \right ),
\end{equation}
for all $i \in \mathcal{A}_1$ and $s \in \mathcal{S}$, which means that in every state $s$, it is better (or equivalent) for player $1$ to employ strategy $\pi^1\left ( s \right )$ than following any pure strategy. \eqref{fundamental} can be expended as 
\begin{equation} \label{V_essence}
\begin{aligned}
&\tilde{r}_{\pi^1}\left ( s \right ) + \gamma\sum_{s'}P_{s\pi^1\left ( s \right )}\left ( s'|s \right )V_{\pi^1}\left ( s' \right ) \\
&\geqslant r\left ( s,i \right )+\gamma\sum_{s'}P_{sa^1=i}\left ( s'|s \right )V_{\pi^1}\left ( s' \right ), \forall i \in \mathcal{A}_1.
\end{aligned}
\end{equation}
\par
In \eqref{V_essence}, note that $P_{s\pi^1\left ( s \right )}\left ( s'|s \right )=g_{\pi}\left ( s' | s \right )$ and $P_{sa^1=i}\left ( s'|s \right )=g_{{\pi}^2|a^1=i}\left ( s'|s \right )$. Expressing the above equation in matrix notation gives 
\begin{equation} \label{V_Q_compare_IRL}
\tilde{r}_{\pi^1} + \gamma G_{\pi}V_{\pi^1}\geqslant r_{a^1=i} +\gamma G_{{\pi}^2|a^1=i}V_{\pi^1}
\end{equation}
where $r_{a^1=i}=C_{a^1=i} r$ and $C_{a^1=i}$ can be constructed from a pure policy where player $1$ will take action $i$ in any state. Substituting \eqref{V_comp2} and \eqref{r_ave_IRL} into \eqref{V_Q_compare_IRL}, gives 
\begin{equation} \label{IRL_constraint}
\left ( F^{\pi^1}_{a^1=i} - C_{a^1=i}  \right ) r \geqslant 0,
\end{equation}
where 
\begin{equation} \label{F}
F^{\pi^1}_{a^1=i} = \left [ \gamma\left ( G_{\pi} - G_{{\pi}^2|a^1=i} \right )\left ( I - \gamma G_{\pi}\right )^{-1} + I  \right ]C_{\pi^1},
\end{equation}
for all $i \in \mathcal{A}_1$.
\par
To establish a Bayesian setting for IRL, we need to assign a prior distribution on the reward vector of player $1$, $f\left ( r \right )$. Let $p\left ( \pi^1|r \right )$ denote the likelihood of observing player $1$'s policy $\pi^1$ when its true reward is $r$. We model $p\left ( \pi^1|r \right )$ by 
\begin{equation}
p\left ( \pi^1| r \right )=\begin{cases}
1, & \mbox{if }Q_{\pi^1}\left ( s,\pi^1\left ( s \right )\right ) \geqslant Q_{\pi^1}\left ( s,i\right ), \forall i \in \mathcal{A}_1\\ 
0, & \mbox{otherwise}.
\end{cases}
\end{equation}
\par
The posterior distribution of the unknown rewards for a given observed
policy $\pi^1$ is now
\begin{equation*}
f\left ( r|\pi^1 \right )\propto p\left ( \pi^1|r \right )f\left ( r \right ),
\end{equation*}
Hence
\begin{equation*}
p\left ( \pi^1| r \right )\propto \begin{cases}
f\left ( r \right ), &  \mbox{if }Q_{\pi^1}\left ( s,\pi^1\left ( s \right )\right ) \geqslant Q_{\pi^1}\left ( s,i\right ), \forall i \in \mathcal{A}_1 \\
0, & \mbox{otherwise}.
\end{cases}	
\end{equation*}
\par
By assuming that $r$ is Gaussian distributed, $r\sim \mathcal{N}\left ( \mu_r,  \Sigma_r\right )$, we can develop a standard optimization program with the posterior of $r$ being the objective function and \eqref{IRL_constraint} being the constraint.  Specifically,
\begin{equation}\label{IRL_complex_program}
\begin{aligned}
\textrm{minimize:}  \quad
& \frac{1}{2}\left ( {r-{\mu_r}} \right )^T {\Sigma_r}^{-1} \left ( {r-{\mu_r}} \right )   \\
\textrm{subject to:} \quad
& \left ( F^{\pi^1}_{a^1=i} - C_{a^1=i}  \right ) r \geqslant 0,
\end{aligned}
\end{equation}
for all $i \in\mathcal{A}_1$. In the above formulation, $\mu_r$ is the mean of the unknown reward vector as a prior, and $\Sigma_r$ is its covariance matrix. 
\section{NUMERICAL EXPERIMENTS}
In this section, we will apply the Bayesian IRL to the abstract soccer game introduced in \cite{Beling2013,Lin2014}, and compare the results with those obtained from MIRL. 
\par
\begin{figure}[h]
  \centering
  \includegraphics[width=3in]{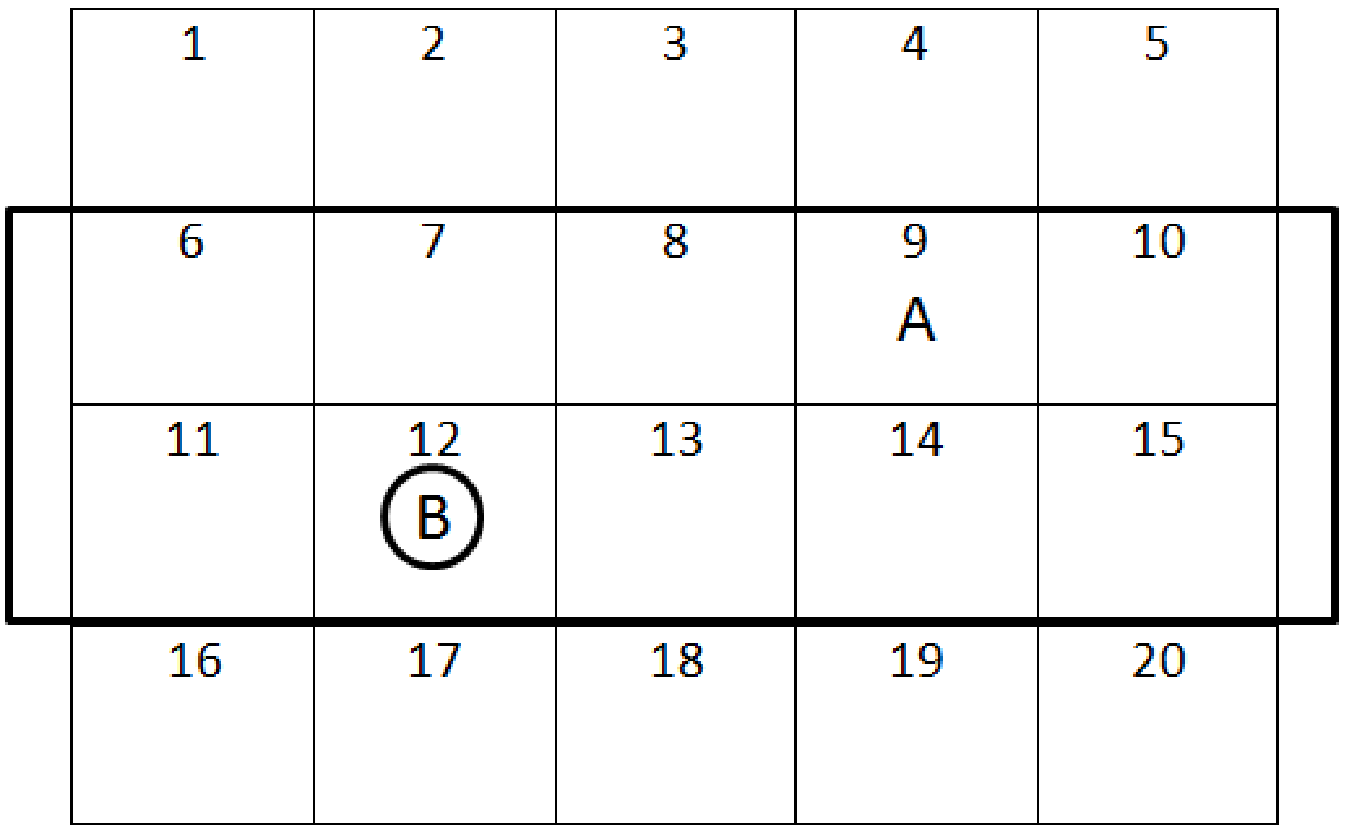}\\
  \caption{Soccer game: initial board}\label{soccer_reset}
\end{figure}

The game is played on a $4 \times 5$ grid as depicted in Figure \ref{soccer_reset}. We use A and B to denote two players, and the circle in the figures to represent the ball. Each player can either stay unmoved or move to one of its neighborhood squares by taking one of 5 actions in each turn: \emph{N} (north), \emph{S} (south), \emph{E} (east), \emph{W} (west), and \emph{stand}. Each player can only take one action in a single time period, and both of them act simultaneously. If both players land on the same square in the same time period, the ball is exchanged between the two players with probability $\beta=0.6$, which is known to the observer. There are in total $800$ states in this model, corresponding to the positions of the players and ball possession. Each player aims to maximize its expected points scored, subject to a discount factor of $\gamma = 0.9$. 
\par
Both players attempt to dribble the ball into specific squares representing their opponent's goal. Player A attempts to score by reaching squares 6 or 11 with the ball, and player B attempts to score by reaching squares 10 or 15. Once a point is scored, the players take the positions shown in Figure \ref{soccer_reset} and ball possession is assigned randomly.
\par
Obviously, the two rational players are playing a zero-sum stochastic game. As stated in Section 3, the bipolicy that they follow is a minimax bipolicy, and is known in this example. We are tasked to recover the reward structure of A, and thereafter infer which squares A must reach in order to score a point (the goal squares). 
\par
\subsection{State Dependent Rewards} \label{chap_simple}
In the above game model, the reward is only state dependent. We will apply MIRL and IRL methods to recover A's rewards. 
\par
We use a Gaussian prior, where the mean reward assigns 0.8 point to player A in every state where A has possession of the ball and -0.8 point in every state where player B has possession of the ball. The covariance matrix of the prior is assumed to be an identity matrix, because without the knowledge of point structure, the correlation between different reward is not clear. 

\begin{figure}[h]
  \centering
  \includegraphics[width=3in]{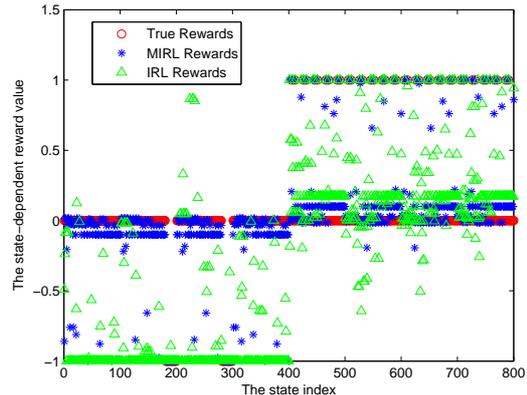}\\
  \caption{State-dependent rewards recovery results}\label{fig:state_dependent}
\end{figure}

Results from these experiments are shown in Figure \ref{fig:state_dependent}.  In the figure, red circles represent the true reward, green triangles represent rewards learned from IRL, and blue stars represent rewards learned from MIRL.  Examination of the figure shows a qualitative advantage for MIRL in that, in aggregate, the blue stars lie substantially closer to the red circles than do the green triangles.  In Section \ref{monte} we assess the quality of learned rewards in terms of the quality of the forward policy that can be learned from them.

\subsection{State and Actions Dependent Rewards}
We now complicate the soccer game by allowing a ``shoot" action. In addition to the available 5 actions, each player who has the ball can take a shot toward their opponent's goal at any position, with a \emph{probability of succesful shot} (PSS) distribution over all positions in the field shown in Table \ref{table:original PSS}. Assume that each one's PSS is independent of its opponent's position. One major difference from the simple model is that now the reward to be covered depends on both state and actions. Recall that IRL is used to infer $r\left ( s, a^1 \right )$ and MIRL will infer $r\left ( s, a^1, a^2 \right )$. We can compare the two results by calculate $r\left ( s, a^1\right )$ from \eqref{reward_transform} in section \ref{sec_IRL}.
\begin{table}
\centering 
\begin{tabular}{c c c c}  
\hline\hline 
  & PSS = 1 & PSS = 0.7 & PSS = 0.5\\  
\hline 
A & 6, 11 & 1, 7, 12, 16 & 2, 8, 13, 17 \\ 
B & 10, 15 & 5, 9, 14, 20 & 4, 8, 13, 19 \\
\hline\hline 
 & PSS = 0.3 & PSS = 0.1 & PSS = 0 \\
 \hline 
A  & 3, 9, 14, 18 & 4, 10, 15, 19 & 5, 20 \\ 
B & 3, 7, 12, 18 & 2, 6, 11, 17 & 1, 16 \\
\hline
\end{tabular}
\caption{Original PSS distribution of each player} 
\label{table:original PSS} 
\end{table}
In both of the methods, we need to assign a mean and a covariance matrix for the prior. We can also develop three types of means based off of our knowledge of this game, as the following:
\begin{itemize}
\item \emph{Weak Mean}: the same as the one described in section \ref{chap_simple};
\item \emph{Median Mean}: guessing that A's goal might be among the rightmost squares, or squares $5$, $10$, $15$ and $20$, and symmetrically, B's goal might be among the leftmost squares, or squares $1$, $6$, $11$ and $16$, we assign $1$ point to A whenever A has the ball and is in the four leftmost squares, and $-1$ point to A whenever B has the ball and is in four rightmost squares. Also, when A has the ball and takes a shot, no matter where she is, we assign $0.5$ point to A. Otherwise, no points will be assigned to A.
\item \emph{Strong Mean}: we have a good guess of A's point distribution, except for its PSS distributions. So comparing to \emph{median mean}, the only difference is that now the potential goal area includes only $2$ squares (square $6$ and $11$ for A and square $10$ and $15$ for B), rather than $4$ squares, for both players.
\end{itemize}
\par
The covariance matrix of the reward vector encodes our belief of the structure of the prior. We can come up with a covariance matrix encapsulating some internal information subject to our knowledge of the relationship between rewards,
\begin{enumerate}
\item When A has the ball and takes a shot, the PSS depends only on A' s position in the field. 
\item In any state when A has the ball, the reward for A for any non-{\em shoot} action is a state-dependent constant.  
\end{enumerate}
We name this covariance matrix \emph{Strong Covariance Matrix}, in order to distinguish it from the simple identity matrix we used in the simple game model. 
\par
\begin{figure}[h]
  \centering
  \includegraphics[width=3in]{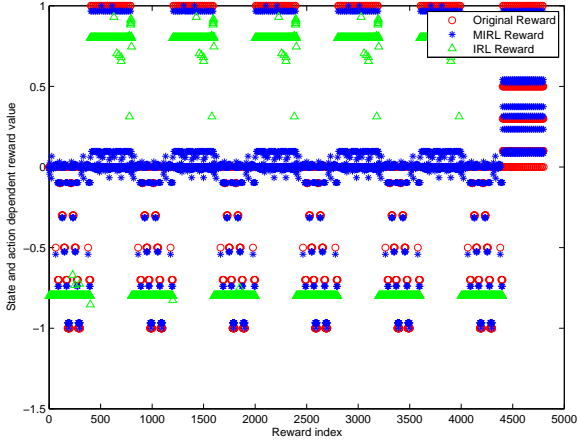}\\
  \caption{MIRL vs IRL on rewards: weak mean and strong covariance}\label{IRL_weakmean_strongcov}
\end{figure}
\begin{figure}[h]
  \centering
  \includegraphics[width=3in]{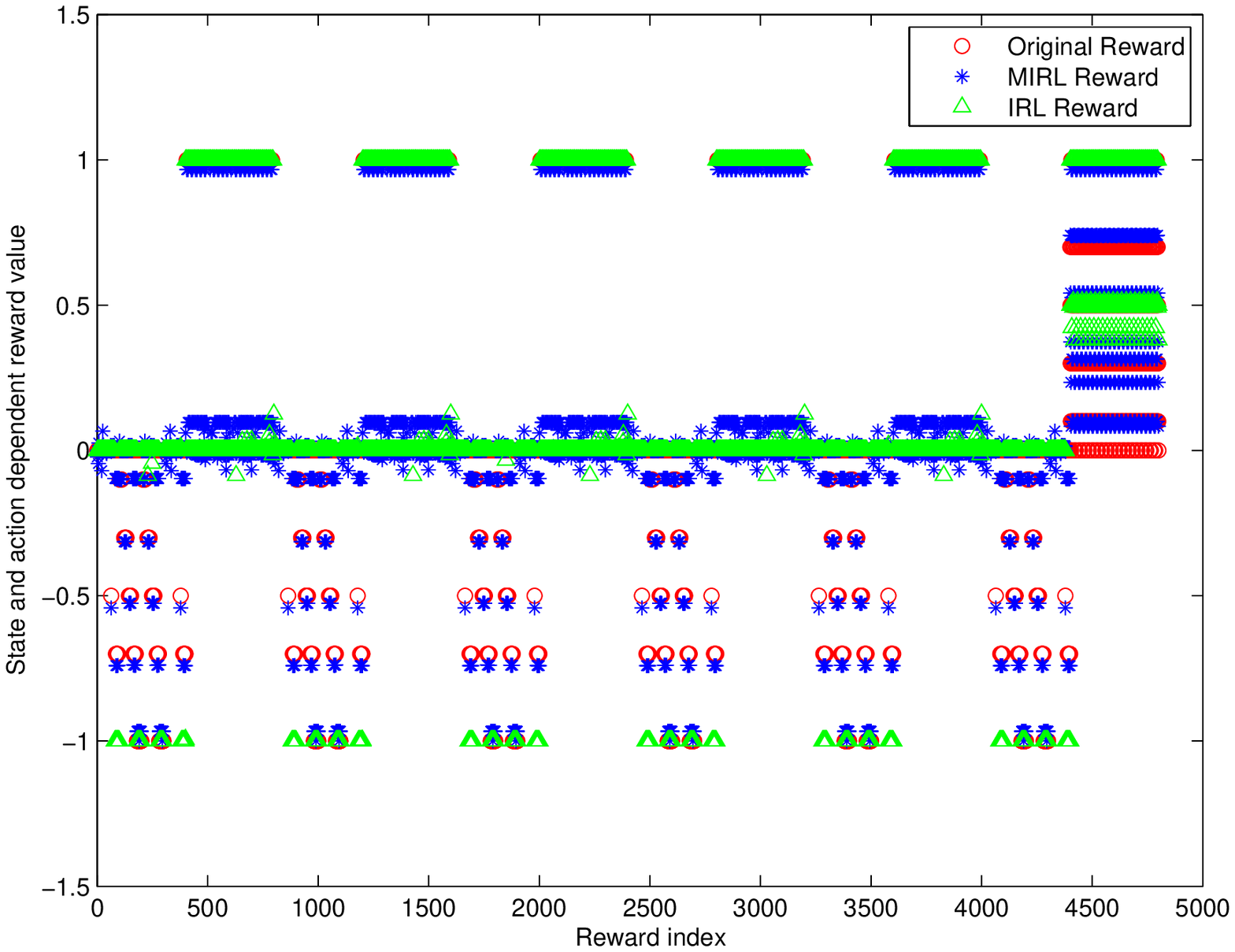}\\
  \caption{MIRL vs IRL on rewards: median mean and strong covariance}\label{IRL_medianmean_strongcov}
\end{figure}
\begin{figure}[h]
  \centering
  \includegraphics[width=3in]{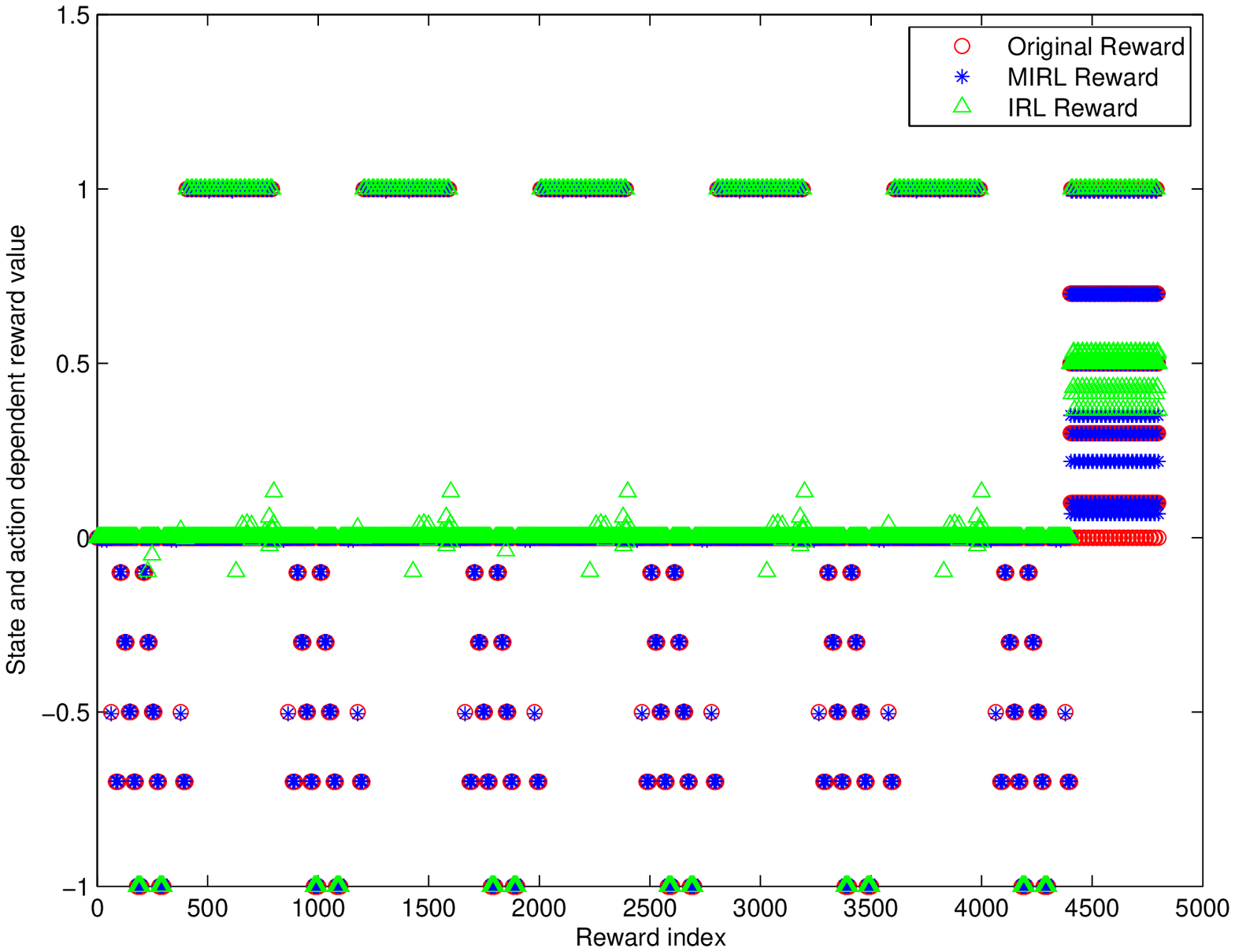}\\
  \caption{MIRL vs IRL on rewards: strong mean and strong covariance}\label{IRL_strongmean_strongcov}
\end{figure}

Figures \ref{IRL_weakmean_strongcov}-\ref{IRL_strongmean_strongcov} show, for the various experiments, original rewards (red circles), rewards learned from IRL (green triangles), and rewards learned from MIRL (blue stars). It can be seen that in each figure there is a considerable overlap in distribution between the true rewards and MIRL rewards.   The recovered rewards from IRL, by contrast, tend to lie far away from the true rewards. 
\par
\begin{figure}[h]
  \centering
  \includegraphics[width=3in]{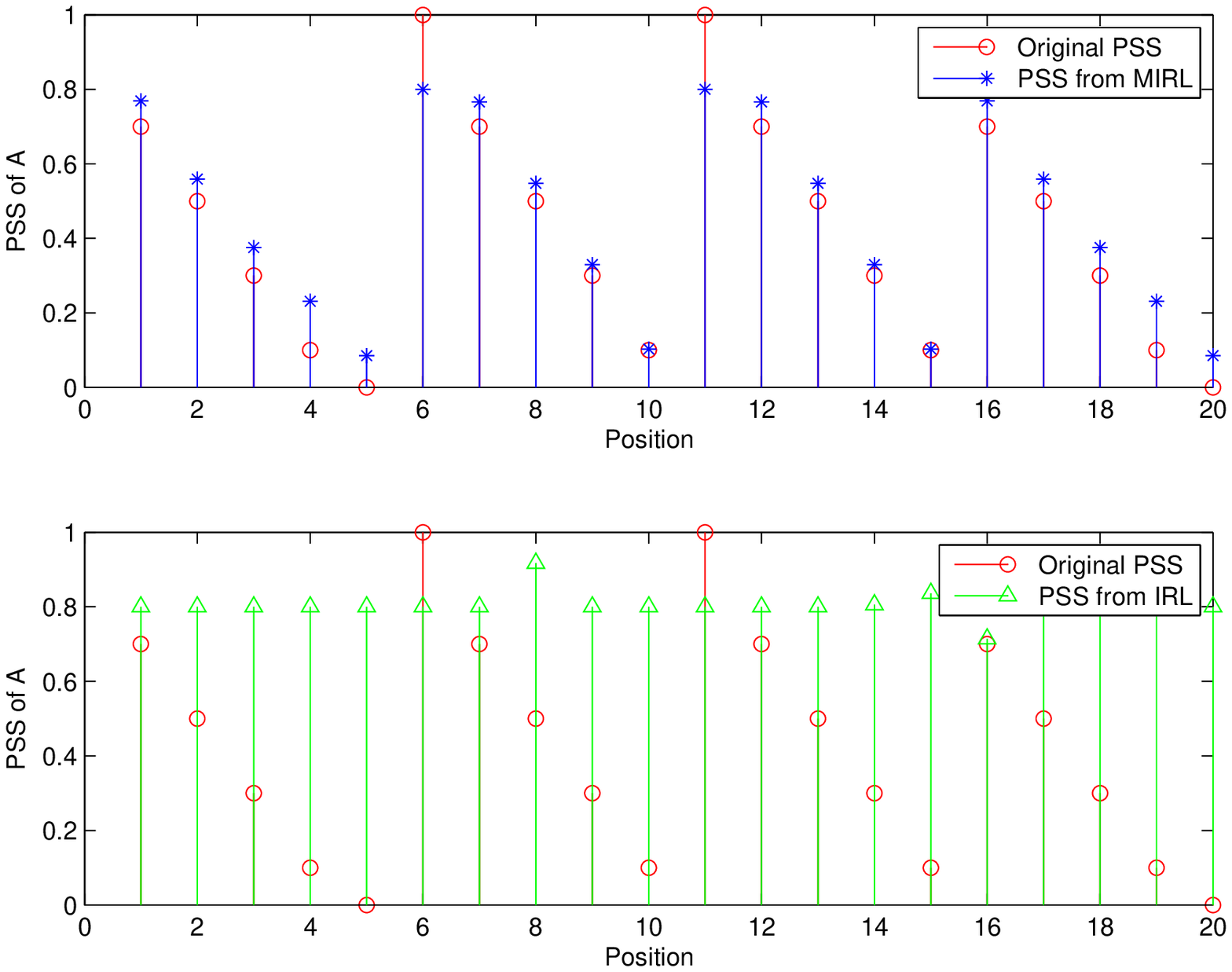}\\
  \caption{MIRL vs IRL on PSS: weak mean and strong covariance}\label{PSS_weakmean_strongcov}
\end{figure}
\begin{figure}[h]
  \centering
  \includegraphics[width=3in]{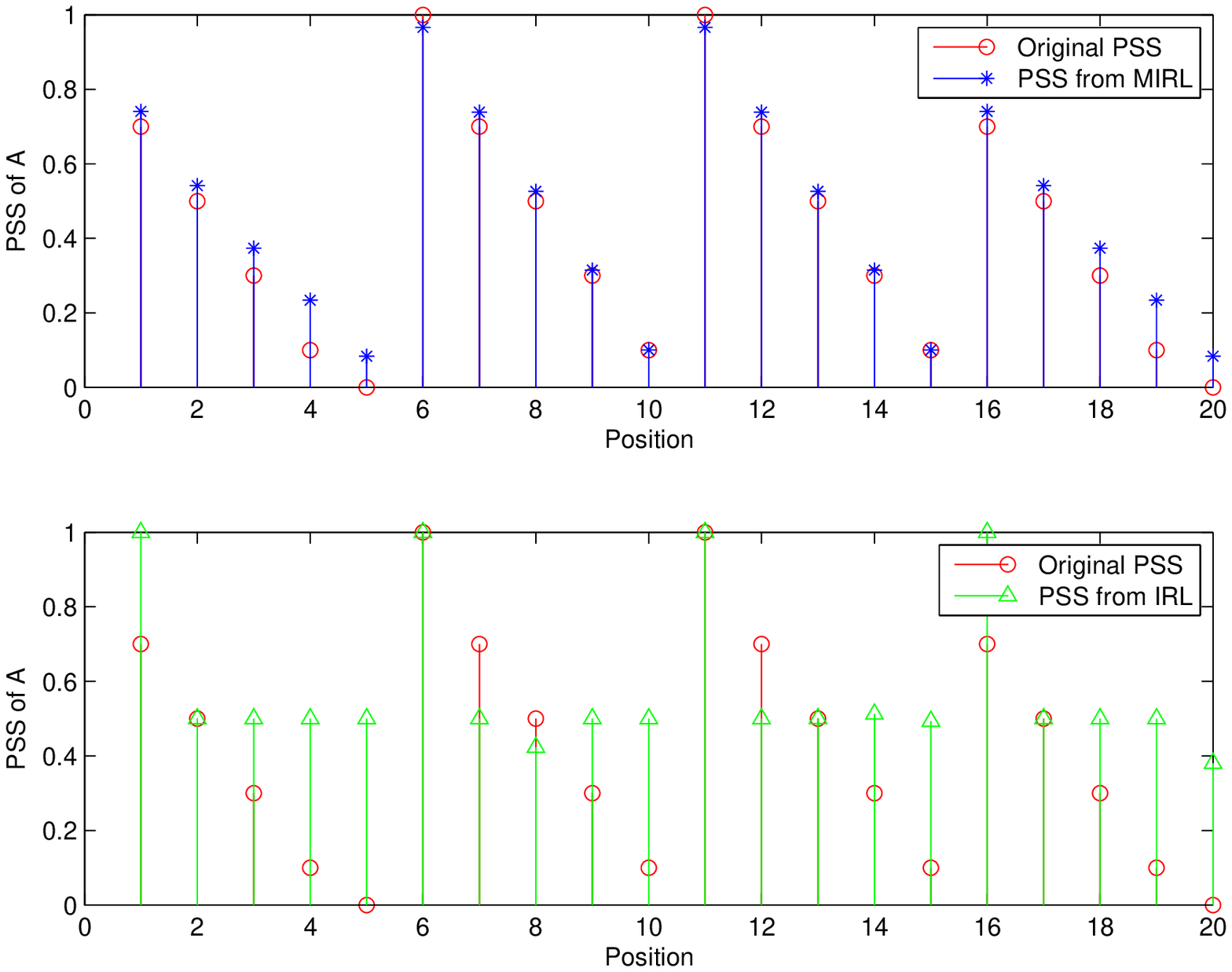}\\
  \caption{MIRL vs IRL on PSS: median mean and strong covariance}\label{PSS_medianmean_strongcov}
\end{figure}
\begin{figure}[h]
  \centering
  \includegraphics[width=3in]{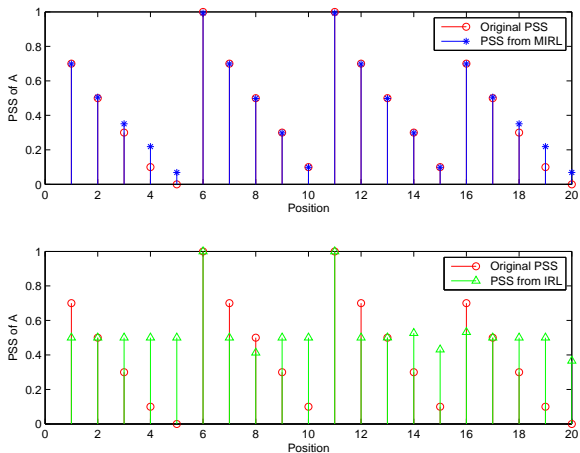}\\
  \caption{MIRL vs IRL on PSS: strong mean and strong covariance}\label{PSS_strongmean_strongcov}
\end{figure}

We also check the recovery of A's PSS, present results in Figure \ref{PSS_weakmean_strongcov}-\ref{PSS_strongmean_strongcov}.
All these results show that compared to the MIRL approach, the IRL method is unable to give a reasonable recovery of A's PSS. 
\section{MONTE CARLO SIMULATION USING RECOVERED REWARDS}
\label{monte}
 In this section, we measure the quality of learned rewards in terms of the quality of the forward solution that they induce. Let A employ the IRL rewards and B employ the MIRL rewards, and both believe that their own reward function is the true one. Being rational, both of them will employ a minimax bipolicy based off of their own rewards. Another criteria to evaluate the rewards quality is to apply them in a different environmental setting, e.g. $\beta=1$. We will simulate games between A and B when $\beta=0.6$ and $\beta=1$, and compare the win-lose results of cases where different sets of rewards are employed.  
\par
The simulation results are presented in Table~\ref{table:simulation_results_beta_04}. In this table, the first column is the type of rewards that A and B employ to develop their minimax policies. A comparison only occurs when the two players use the same type of rewards on their own. \emph{WM}, \emph{MM}, \emph{SM} and \emph{SC} stand for \emph{weak mean}, \emph{median mean}, \emph{strong mean}, and \emph{strong covariance matrix}, respectively. The rest columns are the simulation results of 5000 rounds of games between A and B in cases where $\beta$ being $0$, $0.6$ and $1$. For a more clear comparison, we only count those game episodes ending in win-lose outcomes. Each column gives the winning percentage of B in each different rewards set they use. Our description of the game model indicates that A and B are supposed to be equal in match. However, this simulation results shows that B gets a big edge on A. the We can conclude from that rewards learned from MIRL beat those learned from IRL in quality. 
\begin{table}
\centering 
\begin{tabular}{l c c c} 
\hline\hline 
 & \% won ($\beta = 0$) & \% won ($\beta = 0.6$) & \% won ($\beta = 1$) \\ [0.5ex] 
\hline 
WM \& SC & 100  & 100 & 100 \\ 
MM \& SC & 77.13 & 63.23  & 62.30 \\ 
SM \& SC & 100  & 100 & 100 \\ 
\hline 
\end{tabular}
\caption{A vs B games simulation results } 
\label{table:simulation_results_beta_04} 
\end{table}
\section{CONCLUSION}
The experimental results presented in this paper suggest that the MIRL problem is worth additional study because learned MIRL rewards tend to substantially closer to true rewards and to yield better forward policies than those learned from IRL. Several factors may underlie the performance of MIRL. First, a multi-agent system often involves games while IRL assumes the other agents in the environment are passive. Second, from the perspective of game theory, optimal strategies are generally mixed, and these in turn are difficult to handle in IRL.  Lastly, IRL cannot fully capture equilibrium information, though some equilibrium information can been reflected in the state transition dynamics. 

\ack We would like to thank the financial support for this project from Science Applications International Corporation (SAIC) through the Research Scholars Fellowships Program.

\bibliography{MIRL_database_init}
\end{document}